# BOOTSTRAPPING METHOD FOR DEVELOPING PART-OF-SPEECH TAGGED CORPUS IN LOW RESOURCE LANGUAGES TAGSET- A FOCUS ON AN AFRICAN IGBO

Onyenwe Ikechukwu E[1], Onyedinma Ebele G[1], Aniegwu Godwin E[2] and Ezeani Ignatius M[3]

[1]Department of Computer Science, Nnamdi Azikiwe University, Awka, Nigeria
{ ie.onyenwe, eg.osita}@unizik.edu.ng
[2]Federal College of Education (Technical), Umunze, Nigeria
aniegwuge@gmail.com
[3]University of Sheffield, United Kingdom
ignatius.ezeani@sheffield.ac.uk

*ABSTRACT*

*Most languages, especially in Africa, have fewer or no established part-of-speech (POS) tagged corpus. However, POS tagged corpus is essential for natural language processing (NLP) to support advanced researches such as machine translation, speech recognition, etc. Even in cases where there is no POS tagged corpus, there are some languages for which parallel texts are available online. The task of POS tagging a new language corpus with a new tagset usually face a bootstrapping problem at the initial stages of the annotation process. The unavailability of automatic taggers to help the human annotator makes the annotation process to appear infeasible to quickly produce adequate amounts of POS tagged corpus for advanced NLP research and training the taggers. In this paper, we demonstrate the efficacy of a POS annotation method that employed the services of two automatic approaches to assist POS tagged corpus creation for a novel language in NLP. The two approaches are cross-lingual and monolingual POS tags projection. We used cross-lingual to automatically create an initial 'errorful' tagged corpus for a target language via word-alignment. The resources for creating this are derived from a source language rich in NLP resources. A monolingual method is applied to clean the induce noise via an alignment process and to transform the source language tags to the target language tags. We used English and Igbo as our case study. This is possible because there are parallel texts that exist between English and Igbo, and the source language English has available NLP resources. The results of the experiment show a steady improvement in accuracy and rate of tags transformation with score ranges of 6.13% to 83.79% and 8.67% to 98.37% respectively. The rate of tags transformation evaluates the rate at which source language tags are translated to target language tags.*

## 1. INTRODUCTION

Part of speech (henceforth POS) tagging is the process of assigning a POS or other lexical class marker to each word in a language texts according to their respective POS label according to its definition and context [9]. It is an important enabling task for NLP applications such as a pre processing step to syntactic parsing, information extraction and retrieval, statistical machine translation, corpus linguistics, etc.

In POS tagging experiments, manual POS tagged corpora are required as training data for taggers. Even in some cases, such as unsupervised methods, some manually annotated corpora are still required as a benchmark in evaluating taggers performance. A corpus that is POS untagged in a natural language text is often not suitable for training purposes. Typically, training data are to be POS tagged in some fashion in order to be able to extract and learn the salient textual features. Unfortunately, annotated corpora are in short supply for most low- resource languages such as African languages. Manually annotating even a small corpus is incredibly tedious, time-consuming, costly and is infeasible for larger ones. Automatically POS tagging a corpus with the necessary information is usually not possible since the annotations needed for training are typically the information we are trying to find in the first place [4].

When POS annotating a language corpus with a new tagset, the initial stages of the annotation process face a bootstrapping problem. Since there are no automatic taggers available to help the annotator, the annotation process becomes too laborious to quickly produce adequate amounts of POS tagged corpus for training the taggers. Previous work on bootstrapping has proposed a number of solutions which could be monolingual-based focusing only on the target language, bilingual-based focusing on two languages, and multilingual-based focusing on more than two languages or combination of them.

An obvious example of monolingual-based is to manually annotate a small part of a text, use the annotated text to train a POS tagger. Select a part of the same text that is not annotated (recommended selecting a larger size), use the tagger to annotate the selected part. Hand correct those annotations, retrain a tagger on that corrected part plus the initial one, and so on. Do and continue on this process until a large amount of good quality annotated data is produced. Apart from this example, there are others that use semi-automatic techniques with human annotation expert in the loop [6], [12], [2], [24]. [5] suggested the use of an existing tagger, and devise mapping rules between the old and the new tagset. However, as the construction of such mapping rules requires considerable linguistic knowledge engineering, this solution only shifts the problem to a different domain. COMBI-BOOTSTRAP [29] used existing taggers and lexical resources for the annotation of corpora with new tagsets. The existing resources are used as features for a second level machine learning module, that is trained to make the mapping to the new tagset on a very small sample of annotated corpus material.

Cross-lingual annotation projection method that leverages parallel corpora to bootstrap a POS tagging process without significant annotation efforts for a resource-poor language. There are bilingual-based and multilingual-based approaches, in both at least there is a resource-rich language and thus other languages will have numerous borrowings from it via word-alignment [28], [19] and word-embedding [1].

In this paper, we propose approaches for more efficient POS annotation method. There are two possible automatic approaches to assist POS tagged corpus creation for a novel language in NLP.

1) Monolingual POS tags projection

2) Cross-lingual POS tags projection

The first approach involves to manually annotate some texts, train a Part-of-Speech tagger on the text and POS tag newly selected texts (a size larger than the training data). The assigned tags will then be manually corrected. The hope is that it will take less time to correct the errors in the output than to tag the same material from scratch. The corrected material can be added to the existing training set, and tagger retrained. With each iteration, tagger accuracy should improve and so the effort required for correction reduce. This looks fine since it focuses on the target language and the tagset is correct. The second approach uses automatic alignment between parallel texts as a basis for projecting POS tags from one language to the other. Its completely automatic approach for generating tagged corpus and possible where there are parallel texts and source language has available NLP resources, it becomes easier to automatic project the source language linguistic features onto the target language via automatically word-aligned words, that is, the Yarowasky baseline projection [28]. There are possible limitations to this method, viz; poor accuracy of the word-alignment between the languages involved, inconsistency in the word matching pattern between the two sides of the bilingual text due to translation shortfalls, not all the POS tags desirable in one language can be mapped to the other. Initials tags are solely from the source language and may be the opposite of the target language since the two languages are different. So, if one wants correctly annotated material for a target language, it is not obvious that correcting projected tags is "better" or "close-to" compared to standard manual annotation?

Of the two preceding approaches, the monolingual method looks to be superior, as it produces material for correction that is labelled with the tagset of the target language. The new approach combines 1 and 2 preceding approaches and in two combined methods. It requires few manually annotated texts for the target language, trained tagger on that portion, and the projected tags. The projected initials tags, solely from the source language, will be translated to the target language tagset and the noise induced in the projected data via word-alignment will be cleaned using a rule-based machine learning technique.

The positive effects of this approach are demonstrated in the remainder of this paper, which is sectioned as follows. Section 2 describes the experimental resources that are used in the experiments. Section 3 describes the experiments. Section 4 presents the results of our experiments. Finally, Section 5 presents the summarization and conclusion.

## 2. EXPERIMENTAL RESOURCES

The following sections describe the data and tools we used in this paper.

### 2.1. Data

New World Translation (NWT) Bible [17] provides an ideal test case for our study because of the existence of publically accessible texts of English and their translations in electronic format available in Igbo that is already POS annotated in [23]. The former allows us to use existing English POS tagger on the English texts and transfer POS tags via alignment and projection onto Igbo texts; the latter allows us to evaluate the tagged corpus on sizeable human-annotated tags.

### 2.2. Igbo Language

Igbo is one of the under-resourced languages of the African continent. It is the native language for a subset of Nigerians called Igbo who live in the eastern part of the country. It is a Kwa subgroup language of the Niger-Congo family [11], and one of the most spoken languages of West Africa [13] with its speakers forming about 3% of African and 18% of Nigerian populations [8].

**2.3. POS Tagger Selection and Implementation**

For the purpose of POS tagging the English texts, we chose English Stanford Log-linear Tagger [27], tagging tools that are commonly used and have done well on tagging generally.
FnTBL is a Transformation-based learning in the fast lane based on Brill's TBL [7] reimplemented by [18]. FnTBL provides a platform that enables the use of three input layers that are helpful in this study to reduce noise in a noisy dataset and to translate one state Y to another state X based on contexts. From the TBL documentations [10], the general algorithm description is as follows:

- For presentation clearer, $X$ denotes the sample space, $C$ denotes set of possible classifications of the samples, S denotes the state space, π denotes a predicate defined on the space $S^+$.

- A rule $r$ is defined on a pair $(\pi,c)$ of a predicate and a target state, and it will be written as $\pi=c$. A rule $r=(\pi,c)$ is said to apply on a sample $s$ if the predicate $\pi$ returns true on the sample $s$.

- Given a state $s=(x,c)$ and a rule $r=(\pi,c')$, the state resulting from applying rule $r$ to state $s$, $r(s)$, is defined as $r(s) = \begin{cases} S & \text{if } \pi(s) = \text{false} \\ (x, c') & \text{if } \pi(s) = \text{true} \end{cases}$

- We assume that there are some labeled training data $T$ and some test data $T$ on which to compute performance evaluations.

- –The score associated with a rule $r = (\pi, c)$ is usually the difference in performance (on the training data) that results from applying the rule:

$$Score\ (r) = \sum_{s \in T} score(r(s)) - \sum_{s \in T} score(s)$$

where $score\ ((x, c)) = \{\ldots\ldots$ if $c \neq truth(x)$

Given the above description, TBL algorithm can be now described as follows:

1. Initialize each sample in the training data with a classification (e.g. for each sample x determine its most likely classification c); let T 0 be the starting training data.

2. Considering all the transformations (rules) r to the training data T k , select the one with the highest score

3. Score $(r)$ and apply it to the training data to obtain $T_{k+1} = r\ (T_k) = \{r\ (s)\ |\ s \in T_k\}$. If there are no more possible transformations, or Score $(r) \leq \theta$, stop.

4. $k \leftarrow k + 1$

5. Repeat from step 2.

**2.4. Tagged Igbo and English Parallel Corpora**

We used POS tagged Igbo corpora and the parallel English texts for the purposes of word-alignment, resource borrowings and performance evaluations. Tagged Igbo Corpora (IgbTC) was produced in [23], [24], [21], [22]. It has nearly 300,000 annotated tokens in total and contains 67 tags. The corpus contains several text styles such as essay, news, poem, story, novel, and religious writings. The use of Bible in bootstrapping POS annotation has been widely recognised in the literature by several authors [25], [26], [15], [14], [16], [6], [3] as valuable. The Bible offers some advantages beyond its availability. All its translations are carefully numbered and well structured (books, chapters, verses), facilitating good sentence/word alignments. We took only the religious genre of IgbTC, the New Testament part of New World Translation (NWT) Bible 6, that is parallel translated from the corresponding English version. Thus, we refer to this religious genre as Igbo tagged New Testament texts (*IgbTNT*). Moreover, we POS tagged the corresponding English parallel texts using a Stanford Log-linear English POS Tagger (SLT) [27].

## 3. EXPERIMENT

There are 4 serial experiments conducted in this section, viz; preprocessing of the parallel data, POS tags projection comprising cross-lingual and monolingual experiments for tags transformation and noise reduction, and finally, accuracy evaluations.

### 3.1. Preprocessing of the English and Igbo Parallel Data

The parallel texts of the data we reported in section 2.1 were cleaned, processed and then examined for levels of correspondence for the cross-lingual POS tags projection experiments. The two texts were compared to verify if there were gaps in the chapters and whether one version had more chapters over the other with the aim of removing the non-corresponding chapters. In addition to this, we assume sentences are already verse-aligned, and discrepancies in verses between the two Bibles were corrected. We maintain a sentence length not > 100 for all the texts. The sentence or verse lengths that are more than 100 are split into newlines and identity numbers are assigned to them as links to the original sentences or verse. We are careful not to split a sentence out of context. The verse at line 1780 (last chapter of the book Mak) of the New Testament Bible was 145-word length, it was split into two verses (1780a and 1780b). Furthermore, tokens in this form Mid'i·an, Ca'naan·ites, Am'or·ites, etc., in English and Igbo were normalized to Midian, Canaanites, Amorites. Some tokens like bú instead of bú. , m instead m̀, combining grave accent (.), combining acute accent ( ́ ) and combining dot below (.) that were seen as separate tokens were all corrected. Some of the Hebrew symbols א, i, k associated with the book of Psalms "Abu. O.ma" were removed. This processing resulted in a parallel corpus of 27 books and 8219 verses with 236331 and 263856 words for the English and Igbo New Testament part of NWT Bibles respectively. Note that Igbo part of this parallel corpus is the same as the texts of IgbTNT reported in section 2.4.

### 3.2. POS Tags Projections

Here, we combine two bootstrapping approaches. Firstly, we perform a cross-lingual tag projection experiment on the parallel corpus (from section 3.1) to give us an initial annotated data. Then, we robustly apply monolingual tag projection method for POS tags transformation and noise reduction in the initial annotated data. The overall algorithm for cross-lingual and monolingual tag projections, error correction and noise trimming is illustrated in Figure 1.

#### 3.2.1. Cross-lingual POS Tags Projection Via Word-Alignment

This approach uses automatic alignment between parallel texts as a basis for projecting POS tags from one language to the other. Its completely automatic approach for generating tagged corpus and possible where there are parallel texts and source language has available NLP resources, it becomes easier to automatic project the source language linguistic features onto the target language via automatically word-aligned words.

Word-alignment was automatically computed on the parallel corpus of section 3.1 using Giza$^{++}$ [20]. The first five verses of the New Testament Bible in the parallel corpus are shown in Figure 2. Figure 3 shows some samples of the alignment result on the first five verses of the New Testament Bible in the parallel corpus.

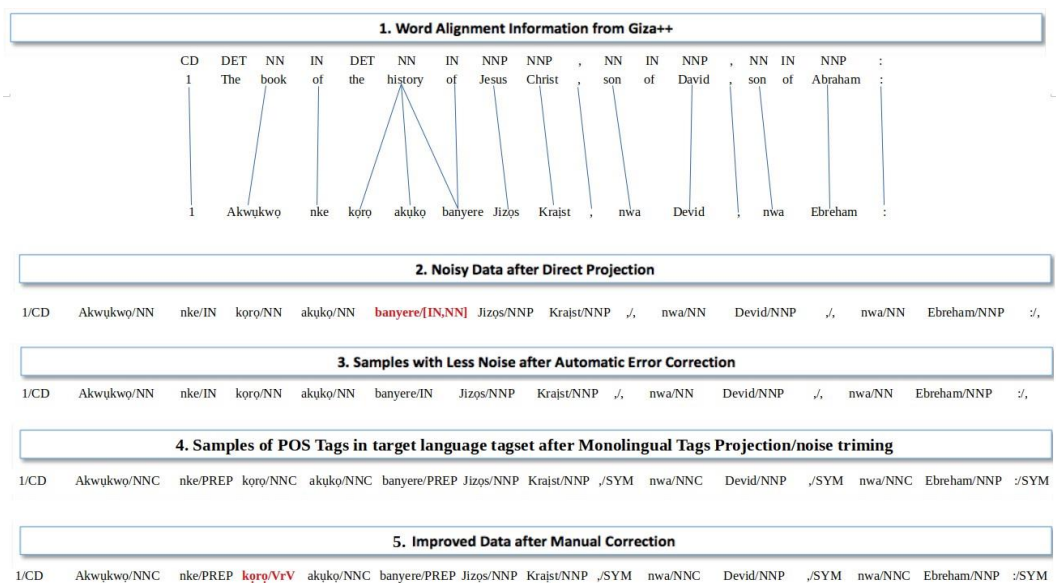

**Figure 1**. The overall algorithm of the POS tags projections methods. See appendix for the meaning of POS tags the target language (Igbo) tagset.

```
1 The book of the history of Jesus Christ , son of David , son of Abraham :
1 Akwụkwọ nke kọrọ akụkọ banyere Jizọs Kraịst , nwa Devid , nwa Ebreham :

2 Abraham became father to Isaac ; Isaac became father to Jacob ; Jacob became father to Judah and his brothers ;
2 Ebreham mụrụ Aịzik ; Aịzik amụọ Jekọb ; Jekọb amụọ Juda na ụmụnne ya ndị ikom ;

3 Judah became father to Perez and to Zerah by Tamar ; Perez became father to Hezron ; Hezron became father to Ram ;
3 Juda amụọ Pirez na Zira , bụ´ndị Tema mụrụ ya ; Pirez amụọ Hezrọn ; Hezrọn amụọ Ram ;

4 Ram became father to Amminadab ; Amminadab became father to Nahshon ; Nahshon became father to Salmon ;
4 Ram amụọ Aminadab ; Aminadab amụọ Nashọn ; Nashọn amụọ Salmọn ;

5 Salmon became father to Boaz by Rahab ; Boaz became father to Obed by Ruth ; Obed became father to Jesse ;
5 Salmọn amụọ Boaz , bụ´onye Rehab mụrụ ya ; Boaz amụọ Obed , bụ´onye Rut mụrụ ya ; Obed amụọ Jesi ;
```

**Figure 2**. The overall algorithm of the POS tags projections methods. See appendix for the meaning of POS tags the target language (Igbo) tagset.

```
1 0-0 1-2 2-3 3-5 4-5 5-5 5-6 6-7 7-8 8-9 9-10 10-12 11-13 12-14 13-16 14-17

2 0-0 1-1 2-2 6-3 3-5 4-6 5-7 6-8 6-9 7-11 8-12 9-13 10-14 10-15 11-17 12-18 14-19 13-20 16-20 17-21

3 0-0 1-1 2-2 2-3 3-5 4-6 5-8 7-8 10-9 8-10 9-10 10-10 11-10 12-11 13-12 14-13 14-14 15-16 16-17 17-18 18-19 18-20 19-22 20-23

4 0-0 1-1 2-2 2-3 3-5 4-6 5-7 6-8 6-9 7-11 8-12 9-13 10-14 10-15 11-17 12-18

5 0-0 1-1 2-2 2-3 5-3 3-5 6-5 7-6 7-7 10-8 11-9 12-10 12-11 15-11 13-13 18-14 8-15 9-15 16-15 17-15 18-15 19-15 20-16 21-17 22-18 22-19 23-21 24-22
```

**Figure 3**. The overall algorithm of the POS tags projections methods. See appendix for the meaning of POS tags the target language (Igbo) tagset.

Comparing Figures 2 and 3, observe the mapping of words between the English and Igbo sentences. For example, **0-0** implies that words at position **0** '*1*' in Igbo ↔ '*1*' in English, **1-2** implies that '*Akwụkwọ*' ↔ '*book*', **2-3** implies that '*nke*' ↔ '*of*', etc. Notice that words do not correspond one-to-one. For example, **5-5**, *5-6* imply that '*banyere*' in Igbo of position **5** is translated in English as '*history*' and '*of*' in positions **5** and **6** (of Figure 3) respectively.

Given the automatically generated alignment in Figure 3, English words' POS tags were automatically projected onto the corresponding Igbo words via the word alignment (see stage 2 of Figure 1). The result of this process is noisy data due to the direct projection of tags. For example, there are 25183 (9.544% of 263856) cases of one-to-many mappings from Igbo to English.

The noisy data (the initial tagged Igbo corpus) where there are multiple alignments are cleaned through automatic error correction. Instances where there n number of tags aligned to a word w/tn (e.g. '*amụọ*' and '*banyere*' in Table 1), we first consider the most frequent tag for that word. Next, where the number of unique alignment tags is equal (e.g. '*banyere*' in Table 1), we calculate the probability of that word given each alignment tag in the entire corpus, then take the tag with the highest probability score. *P(wi/tij)* represents probability, given a tag *tij* aligned to the word *wi* in the set of tags *{ti,1, ti,2, …, ti,n}*, we compute

$$p(w_i | t_{i,j}) = \frac{C(t_{i,j}, w_i)}{C(t_{i,j})}$$

that is how often *ti,1* is associated with *wi*, *ti,2* is associated with *wi* and so on. The result from this process gave projected POS annotated Igbo corpus in English tagset we used in the following experiments. This serves as the initial tagged Igbo corpus (*IgbTC-0*).

**Table 1**. Words with multiple alignments disambiguated by choosing the most frequent tags and probability tag distribution *P(w/t)*. The final result is a full disambiguated initial projected POS tagged Igbo Corpus (IgbTC-0).

| Igbo | English | Alignment Tag | Most Frequent Assigned Tag | Probability Assigned Tag |
|---|---|---|---|---|
| Matiu | Matthew | NNP | - | - |
| 1 | 1 | CD | - | - |
| 1 | 1 | CD | - | - |
| Akwụkwọ | book | NN | - | - |
| nke | of | IN | - | - |
| kọrọ | history | NN | - | - |
| akụkọ | history | NN | - | - |
| banyere | [history,of] | [NN,IN] | - | IN |
| … | … | … | … | … |
| 2 | 2 | CD | - | - |
| … | … | … | … | … |
| Aịzik | Isaac | NNP | - | - |
| amụọ | [father,became,father] | [NN,VBD,NN] | NN | - |

**3.2.2. Monoligual Tags Projection on IgbTC-0**

There are major limitations to the experiment conducted in section 3.2.1, viz; poor accuracy of the word-alignment between the languages involved, not all the POS tags desirable in one language can be mapped to the other, and initials tags are solely from the source language and may be opposite of the target language since the two languages are different. Thus, we robustly apply

International Journal on Natural Language Computing (IJNLC) Vol.8, No.1, February 2019

monolingual tag projection method that translated the POS tags of the source language (English) to the target language (Igbo) tagset, and the noise induced in *IgbTC-0* via word- alignment are cleaned through the illustration in Figure 1.

Resources used in this method are manually annotated Igbo texts to represent Igbo tagset (which is *IgbTC-s*: a small subset of *IgbTNT* in section 2.4), the corresponding English projected tags from *IgbTC-0* and trained tagger on that portion. Our aim is, to begin with transformation-based learning (TBL), a certified machine learning approach good for small amount of data, where the tags of *IgbTC-s* and the corresponding projected English tags from *IgbTC-0* serve as the truth and initial states to which TBL applies. The TBL arrangement (see the algorithm in section 2.3) enables the automatic transformation of English tags to their corresponding Igbo equivalents. From Table 2, observe the three input layers of the model that are word instance and the two alternative labels: an initial label (English projected tags) and a true label (tags from Igbo tagset).

**Table 2**. The format of data given to TBL as input for training. Columns 1 and 2 are the corresponding *IgbTC-s* (columns 1 and 3) from *IgbTC-0*. TBl will learn how to transform column 2 to 3. English tags are those tags projected from English resources onto *IgbTC-0* via alignment conducted in section 3.2.1. See Table in Appendix A for tags meaning.

| Word | Initial State = English Projected tags | Truth State = Igbo Tags |
|---|---|---|
| Matiu | NNP | NNP |
| 1 | CD | CD |
| 1 | CD | CD |
| Akwụkwọ | NN | NNC |
| nke | IN | PREP |
| kọrọ | NN | VrV |
| akụkọ | NN | NNC |
| banyere | IN | PREP |
| Jisọs | NNP | NNP |
| Kraịst | NNP | NNP |

This is because the rule-based tagger we use for this task employs the transformation-based error-driven learning (TBL) algorithm [7]. TBL requires a truth state representation of the data, i.e. showing the correct label for each item. TBL also creates an initial state labelling of the data, typically using a simple method, such as assigning each item its most common label. The initial state will contain many errors – we override this phase of TBL using the process in section 3.2.1, which is data in Table 2 columns 1 and 2. TBL then proceeds to learn a series of transformation rules, that correct errors in the initial state, so that it better approximates the truth state. These rules are context-dependent, i.e. can apply to replace label *X* with *Y* provided the context meets some requirement, e.g. that the item to the left is some specific *w* or the label to the right is some specific *t*. At run-time, TBL labels unseen data by creating its initial state and then applying the sequence of transformation rules learned during training.

To illustrate the above process, we used *IgbTNT* (the New Testament part of *IgbTC*) in section 2.4 which is the same as Igbo texts in the parallel corpus of section 3.1. We took 5% of *IgbTNT* to be *IgbTC-s* and the corresponding English projected tags from *IgbTC-0* to represent training data on Table 2, train FnTBL (a TBL Part-of-Speech tagger) on the text and apply it on *IgbTC-* Then we randomly took another 5% of *IgbTNT*, add it to the first 5% (now 10%), apply FnTBL on it, and so on. This process follows iterative steps and the algorithm is further illustrated as:

1. take 5% of *IgbTNT* and the corresponding English projected tags in *IgbTC-0* torepresent FnTBL (a TBL Part-of-Speech tagger) training data on Table 2. This 5% represents the manually corrected version of the 5% we are to select and correct from *IgbTC-0*. The POS tags of *IgbTNT* represents the truth state while POS tags from *IgbTC-0* represents the 'errorful' initial state of FnTBL.

2. train FnTBL on 1.

3. apply 2 on *IgbTC-0*, the 'errorful' initial state to transform *IgbTC-0* state to *IgbTC-i*, where i is the number of the current iteration. Our aim is to transform *IgbTC-0* to *IgbTC-1*, *IgbTC-2*, ... until it resembles or 'close to' *IgbTNT*. See Figures 4 and 5.

4. take another 5%, add it to 1 and repeat from 2.

The idea is that it will take less time to correct the errors in the output (see Figure 1) than to tag the same material from scratch. The corrected material can be added to the existing training set, and tagger retrained. With each iteration, tagger accuracy should improve and so the effort required for correction reduce. This looks fine since it focuses on the target language and the tagset is correct. Again, this overrides the idea of constructing mapping rules or lookup dictionary table between the source and target languages tagsets, since the construction requires considerable linguistic knowledge engineering.

## 4. DISCUSSION

In the above experiment, we performed 10 iterative steps in the cleaning of errors and transforming of English POS tags in *IgbTC-0* (produced in section 3.2.1) to their Igbo equivalents. The implication here is that *IgbTC-0* went through 10 states of *tag transformation* and *error cleansing* from *IgbTC-0* to *IgbTC-10*. There are two results we discuss in this section, viz; *tagging accuracy* and *rate of tag transformation* from English tagset to Igbo tagset. These were measured using IgbTNT as the standard against which all the transformation states of *IgbTC-0* are compared. Accuracy was computed using metrics for evaluating POS tagging systems:

$$accuracy = \frac{number\ of\ words\ correctly\ tagged\ by\ tagger}{total\ number\ of\ words}$$

$$errorrate = 1.0 - accuracy$$

The following Figures 4 and 5 present the experimental results of each state of transforming *IgbTC-0* to its truth state, *IgbTNT* (see the algorithm in section 3.2.2). Table 3 presents the scores of the data points in Figures 4 and 5. From Figures 4, we observe an accuracy range of 6.13% to 83.79%. There is a huge jump from states *IgbTC-0* to *IgbTC-1*. *IgbTC-0* is the 'errorful' initial state developed in section 3.2.1 on which FnTBL trained on the first 5% of *IgbTNT* applied to get *IgbTC-1*. FnTBL trained on another randomly selected 5% plus the first 5% of *IgbTNT* was applied on *IgbTC-1* to yield *IgbTC-2* state. Observe a steady and moderate increase in the accuracy from *IgbTC-1* to *IgbTC-10*. The last state accuracy result of 83.79% shows that there are 16.21% errors in IgbTC-10 which will take less time to correct than to tag the same material from scratch.

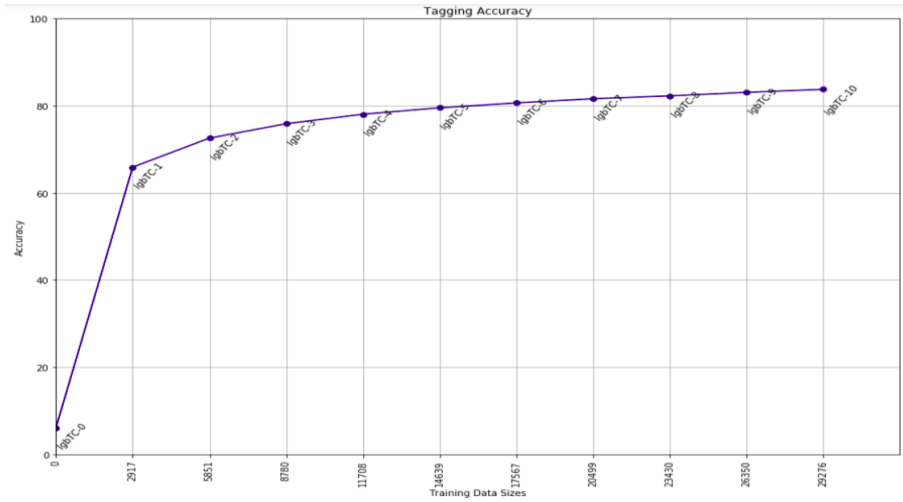

**Figure 4**. Part-of-Speech tagging accuracies of FnTBL tagger at each iteration of *IgbTC-0* transformation process in section 3.2.2

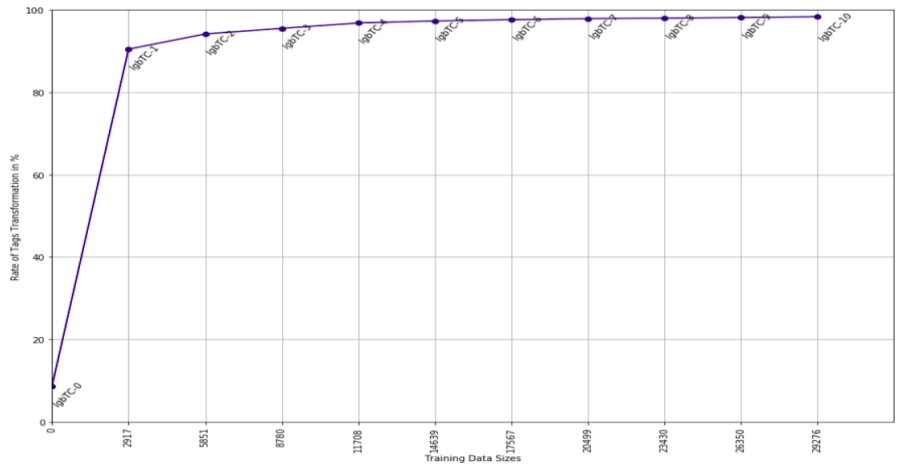

**Figure 5**. The rate at which source language tags (English) is translated to target language tagset (Igbo) in *IgbTC-0* following the algorithm described in section 3.2.2

Table 3. Tagging accuracy and rate of tags transformation from English tags to Igbo tagset starting on *IgbTC-0* to *IgbTC-10*.

| Transformation states of IgbTC-0 | Tagging Accuracy | Rate of Tags Transfroming from Eng Tags to Igbo tags |
|---|---|---|
| IgbTC-0 | 06.13% | 08.67% |
| IgbTC-1 | 65.92% | 90.49% |
| IgbTC-2 | 72.59% | 94.17% |
| IgbTC-3 | 75.89% | 95.54% |
| IgbTC-4 | 78.08% | 96.84% |
| IgbTC-5 | 79.55% | 97.34% |
| IgbTC-6 | 80.67% | 97.64% |
| IgbTC-7 | 81.64% | 97.91% |
| IgbTC-8 | 82.29% | 98.01% |
| IgbTC-9 | 83.09% | 98.18% |
| IgbTC-10 | 83.79% | 98.37% |

At the transformation rate computation, since we are changing the source language (English) tagset in *IgbTC-0* to target language (Igbo) tagset, we look at how many tags of those words in *IgbTC-0* are correctly transformed at each stage of the experiment (see section 3.2.2). We compute this at each stage of the experiment by **dividing** *the number of words whose initial tags have been correctly re-labelled with the tags of the target language tagset* **with** *the total number of words*. *IgbTC-0* at zero state –just as is after cross-lingual POS tagging via alignment in section 3.2.1– contains two POS tags (NNP and CD) that have the same representation and meaning as in Igbo. Thus, Figure 5 shows that these tags form 8.67% of *IgbTC-0* at that initial state. Interestingly, there is a substantial transformation rate from states *IgbTC-0* to *IgbTC-1* just as is the case in the tagging accuracy above. The rate of transformation scores are shown in Table 3. The last state score, 98.37%, shows that there are 1.63% English tags still remaining in the *IgbTC-10*, which are easily corrected.

## 5. CONCLUSIONS

Part-of-speech tagging of a new language corpus with a new tagset at the initial POS tagging stages usually face a bootstrapping problem. Since there are no automatic taggers or other Natural Language Processing tools available in the language to help the annotators, the annotation process becomes too laborious to quickly produce adequate amounts of POS tagged corpus for training the taggers. In this paper, we have demonstrated the efficacy of a more efficient POS annotation method that employed the services of two automatic approaches to assist POS tagged corpus creation for a novel language in NLP.

The two approaches are cross-lingual and monolingual POS tags projection. In the first approach, we used automatic alignment to develop a parallel corpus we used as a basis for projecting POS tags from rich resource source language to the low resource target language We used English and Igbo in this paper. This is possible because there are parallel texts that exist between English and Igbo, and the source language English has available NLP resources. The English language POS tags were automatically transferred onto the corresponding Igbo language via automatically word-aligned words. There are some possible limitations to this method, such as poor accuracy of the word-alignment between the languages involve, inconsistency in the word matching pattern between the two sides of the bilingual text due to translation shortfalls, not all the POS tags desirable in one language can be mapped to the other, and initials tags are solely from the source language and may be opposite of the target language since the two languages are different. Hence the second approach, monolingual POS tags projection. In this approach, we used few manually annotated texts, trained a Part-of-Speech tagger on the text, then used the trained tagger to correct errors induced in the projected tagged corpus created by cross-lingual projection via word-alignment and to change source language tags to the target language tags. The idea is that it will take less time to correct the errors in the output than to tag the same material from scratch. The corrected material can be added to the existing training set, and tagger retrained. With each iteration, tagger accuracy should improve and so the effort required for correction reduce. This looks fine since it focuses on the target language and the tagset is correct. Again, this new approach overrides the idea of constructing mapping rules or lookup dictionary table between the source and target languages' tagsets, since the construction requires considerable linguistic knowledge engineering.

We used FnTBL in this experiment since it is a certified machine learning approach good for small amount of data, where the Igbo tags in the manually annotated Igbo texts (IgbTC-s) and the corresponding English tags in the projected tagged corpus (IgbTC-0) serve as the truth and initial states to which FnTBL applied. This arrangement enables us to automatically transformed English tags to their corresponding Igbo equivalents and reduce induced errors via word- alignment. The results of the experiment show a steady improvement in accuracy and rate of tags transformation with score ranges of 6.13% to 83.79% and 8.67% to 98.37% respectively. The rate of tags transformation evaluates the rate at which source language tags are translated to target language tags.

## APPENDIX

**A.** Table 4. Igbo Tagset tags description and usage. See [21], [22], [23] for the full description of Igbo tagset.

| Tag Name | Tagging Description |
|---|---|
| NNP | Proper noun |
| NNC | Common noun |
| NNM | Number Marking Noun for plurality |
| NNQ | Qualificative noun |
| NND | Adverbial noun |
| NNH | Inherent complement noun use for the completion of the sense of a verb |
| NNCV | Multiword noun formed via verb nominalization |
| NNCC | Inherent complement noun of NNCV |
| VIF | Infinitive verb |
| VSI | Simple verb |
| VCO | Compound verb |
| VMO | Modal verb supplemented by modal suffixes |
| VMOV | Modal verb that require inherent complement noun |
| VMOC | Inherent complement noun of VMOV |
| VAX | Auxiliary verb |
| VPP | Participle |
| VCJ | Conjuctional verb |
| BCN | Bound Cognate Noun |
| VGD | Gerund |
| ADJ | Adjective |
| PRN | Pronoun |
| PRNREF | Reflexive pronoun |
| PRNEMP | Emphatic pronoun |
| PRNYNQ | Pronoun Yes/No Question |
| BPRN | Bound pronoun |
| ADV | Adverb |
| CJN | Conjuncion |
| CJN1 | First correlative conjunction |
| CJN2 | Second correlative conjunction |
| PREP | Preposition |
| QTF | Quantifier |
| DEM | Demonstrative |
| INTJ | Interjection |
| FW | Foreign/Borrowed word |
| SYM | Punctuations |
| CD | Numbers |
| WH | Interrogative |
| IDEO | Ideophone |
| LTT | Alphabets/Letters |
| TTL | Title |
| ENC | Collective, adverbial aditive, negative interrogative, adverbial confirmation, adverbial immediate, present and past |
| VrV | Active/Stative verb |

**AUTHORS**

**Ikechukwu E. Onyenwe** holds a B.Sc. and an M.Sc. in Computer Science from Nnamdi Azikiwe University, Nigeria; a PHD in Computer science from the University of Sheffield, UK. He is currently a researcher/lecturer in the department of Computer Science, Nnamdi Azikiwe University, Anambra State, Nigeria. His research interests are AI, Natural Language Processing, Computational Sciences, Machine Learning, Data Science/Communication, Web/Internet Computing.

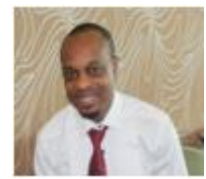

**Onyedinma Ebele G** holds a B.Sc., an M.Sc. and a Ph.D. in Computer science from Nnamdi Azikiwe University, Nigeria. He is currently a researcher and a Senior Lecturer in the Computer Science department of Nnamdi Azikiwe University, Anambra State, Nigeria. His research interests are image processing and AI.

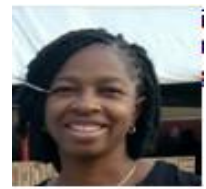

**Aniegwu Godwin E** holds a BSc in Computer science from Nnamdi Azikiwe University, Nigeria; a professional diploma in education (PDE) from Federal College of Education (Technical), Umunze, Nigeria. He is currently a lecturer in the Computer Science Education department of the mentioned college. His research interests are AI, Natural Language Processing and Data Science.

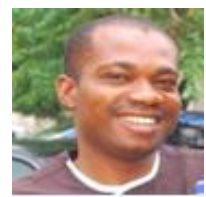

**Ignatius M. Ezeani** holds a BSc in Computer Science from Nnamdi Azikiwe University, Nigeria; an MSc in Advanced Software Engineering from Bournemouth University, UK. Worked as a lecturer and researcher at Nnamdi Azikiwe University, Nigeria from 2003 to 2014. Just concluding Ph.D. at the University of Sheffield, UK. He is currently a Research Associate at Lancaster University, UK. Current research interest is in efficient adaption of existing natural language processing tools and techniques to dealing with the challenges of integrating majority of the low-resource languages in a globalized world for task oriented systems. He is generally interested in the design and development of machine learning and deep neural models and their applications to, not just NLP, but to the broader field of data science.

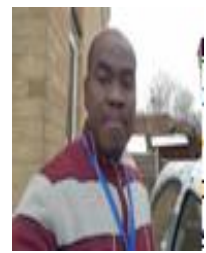